\documentclass[10pt,twocolumn,letterpaper]{article}

\usepackage[accsupp]{axessibility}  % Improves PDF readability for those with disabilities.
\usepackage{iccv}              % To produce the CAMERA-READY version
\usepackage{multirow}
\usepackage[utf8]{inputenc}
\usepackage{mwe}

\usepackage{bm}
\DeclareUnicodeCharacter{394}{$\Delta$}
\usepackage{xcolor}

\usepackage[ruled,linesnumbered]{algorithm2e}

\usepackage{algorithm}
\usepackage{algorithmic}

\usepackage{placeins}

\definecolor{iccvblue}{rgb}{0.21,0.49,0.74}
\usepackage[pagebackref,breaklinks,colorlinks,allcolors=iccvblue]{hyperref}

\def\paperID{4657} % *** Enter the Paper ID here
\def\confName{ICCV}
\def\confYear{2025}

\title{LVFace: Progressive Cluster Optimization for Large Vision Models in Face Recognition}

\author{Jinghan You\thanks{These authors contributed equally.}, Shanglin Li\footnotemark[1], Yuanrui Sun\footnotemark[1], Jiangchuan Wei, \\Mingyu Guo\thanks{Project lead}, Chao Feng\thanks{Corresponding}, Jiao Ran\\
  ByteDance Inc. \\
  \texttt{\{youjinghan, guomingyu.313, chaofeng.zz\}@bytedance.com} \\
}

\begin{document}
\maketitle
\begin{abstract}

Vision Transformers (ViTs) have revolutionized large-scale visual modeling, yet remain underexplored in face recognition (FR) where CNNs still dominate. We identify a critical bottleneck: CNN-inspired training paradigms fail to unlock ViT's potential, leading to suboptimal performance and convergence instability.To address this challenge, we propose LVFace, a ViT-based FR model that integrates Progressive Cluster Optimization (PCO) to achieve superior results. Specifically, PCO sequentially applies negative class sub-sampling (NCS) for robust and fast feature alignment from random initialization, feature expectation penalties for centroid stabilization, performing cluster boundary refinement through full-batch training without NCS constraints. 
LVFace establishes a new state-of-the-art face recognition baseline, surpassing leading approaches such as UniFace and TopoFR across multiple benchmarks. Extensive experiments demonstrate that LVFace delivers consistent performance gains, while exhibiting scalability to large-scale datasets and compatibility with mainstream VLMs and LLMs. Notably, LVFace secured 1st place in the ICCV 2021 Masked Face Recognition (MFR)-Ongoing Challenge (March 2025), proving its efficacy in real-world scenarios. Project is available at \url{https://github.com/bytedance/LVFace}.

\end{abstract}

\section{Introduction}
% The face is a fundamental research topic in computer vision, spanning subfields such as face recognition, reconstruction, and forgery detection, with recognition being a core focus.

% Transformers have revolutionized artificial intelligence, achieving remarkable success in natural language processing through large language models (LLMs) that exhibit consistent performance improvements with increased scale \cite{vaswani2023attentionneed,kaplan2020scalinglawsneurallanguage}. This success has spurred the development of Large Vision Models (LVMs) in computer vision, where Transformers now dominate tasks such as image classification \cite{dosovitskiy2021imageworth16x16words}, object detection \cite{carion2020endtoendobjectdetectiontransformers}, and video processing \cite{zhou2018endtoenddensevideocaptioning}. Unlike CNNs, which rely on local receptive fields, Transformers leverage self-attention mechanisms to model global context, offering superior scalability and effectiveness for complex vision tasks.

% Despite these advancements, face recognition remains predominantly CNN-driven. While recent efforts have explored Transformer architectures \cite{zhong2021facetransformerrecognition, dan2023transfacecalibratingtransformertraining}, two critical challenges persist: (1) the limited scale of face recognition datasets hinders effective Transformer training, and (2) the design of loss functions—crucial for face recognition—remains underexplored in Transformer-based approaches. These limitations suggest that Transformers' full potential in face recognition is yet to be realized.

The human face is a fundamental research topic in computer vision, spanning subfields such as face recognition \cite{deng2019arcface, wang2018cosface, zhao2023cross}, reconstruction\cite{deng2019accurate, li2024uv, lattas2023fitme}, animation \cite{siarohin2019first, zeng2022fnevr, zeng2023face} and anti-spoofing \cite{liu2019deep, liu2024cfpl}, with face recognition (FR) as a core focus. While deep learning has driven significant progress in these areas, the field has been dominated by convolutional neural networks (CNNs). Meanwhile, Transformers have revolutionized artificial intelligence, achieving remarkable success in natural language processing through large language models (LLMs) that exhibit consistent performance improvements with increased scale \cite{vaswani2023attentionneed,kaplan2020scalinglawsneurallanguage}. This success has spurred the development of Large Vision Models (LVMs), where Transformers now dominate tasks such as image classification \cite{dosovitskiy2021imageworth16x16words}, object detection \cite{carion2020endtoendobjectdetectiontransformers}, and video processing \cite{zhou2018endtoenddensevideocaptioning}. Unlike CNNs, which rely on local receptive fields, Transformers leverage self-attention mechanisms to model global context, offering superior scalability and effectiveness for complex vision tasks.

Despite these advancements, face recognition remains predominantly CNN-driven. Although recent efforts have explored Transformer architectures \cite{zhong2021facetransformerrecognition, dan2023transfacecalibratingtransformertraining}, two critical challenges persist: (1) the limited scale of face recognition datasets hinders effective Transformer training, and (2) the design of loss functions—crucial for face recognition—remains underexplored in Transformer-based approaches. These limitations suggest that Transformers' full potential in face recognition is yet to be realized, presenting an important direction for future research.

\begin{figure*}[t]
    \centering
    \includegraphics[width=0.9\linewidth]{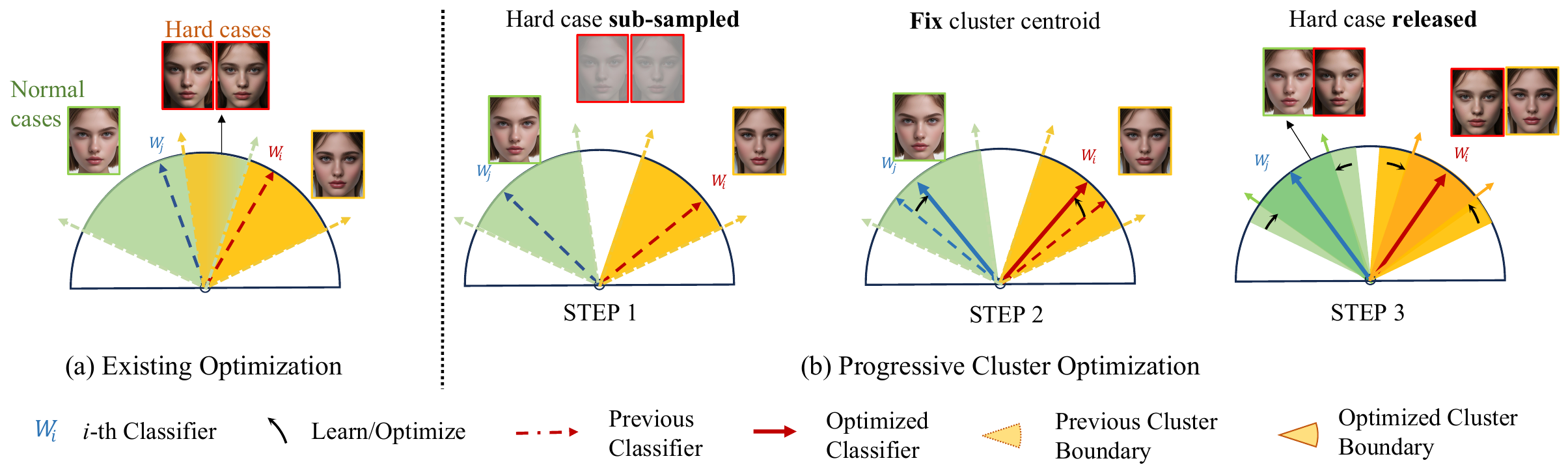}
    \caption{Illustration of our motivation. (a) Conventional one-step optimization struggles with hard cases, leading to ambiguous class boundaries; (b) Our three-stage progressive approach. Stage 1: Hard case sub-sampling for efficient feature alignment; Stage 2: Class centroid stabilization through feature expectation; Stage 3: Cluster boundary refinement via hard case optimization.}
    \label{fig:motivation}
\vspace{-2mm}
\end{figure*}

We observe, as illustrated in \cref{fig:motivation}, existing optimization methods, though effective for small-scale CNN training, struggle to perform as expected in large-scale face recognition scenarios. Inspired by the multi-stage training paradigm of LVMs and LLMs, we propose a step-wise optimization approach that decomposes the learning process into multiple phases, each with explicit optimization targets, to achieve compact and discriminative feature distributions. 

In this work, we propose LVFace, a Transformer-based \textbf{L}arge \textbf{V}ision model for \textbf{Face} recognition, with a novel Progressive Cluster Optimization (PCO) mechanism and a complementary Cosine Stage Scheduler (CSS) . LVFace consists of three stages: (1) \textit{Feature Alignment}, where partial negative sampling and a modified CosFace loss \cite{wang2018cosface} mitigate noise during early-stage feature alignment; (2) \textit{Centroid Stabilization}, which employs feature expectation penalties to anchor cluster centers near normal samples while retaining hard sample learning for robust generalization; and (3) \textit{Boundary Refinement}, where full-sample training refines decision boundaries of each cluster to maximize inter-class margins and minimize intra-class variance. To control transitions between these stages, CSS monitors the cosine similarity between sample features and their class centroids. This ensures that stage transitions occur only when representations exhibit statistically significant improvements in discriminative power.

Experiments on the MFR-Ongoing \cite{deng2021masked}, IJB-B, and IJB-C \cite{maze2018iarpa} benchmarks demonstrate that LVFace outperforms state-of-the-art methods. These results suggest that large-scale datasets and well-designed loss functions can eliminate the need for domain-specific inductive biases, unlocking Transformers' full potential in face recognition.

The main contributions of this paper are as follows:
\begin{itemize}
    \item We propose LVFace, a ViT-based face recognition model that leverages progressive cluster optimization with a cosine stage scheduler to mitigate the challenges of FR optimization in LVMs. LVFace achieves state-of-the-art performance while preserving feature compatibility with mainstream VLMs and LLMs.
    \item We systematically investigate multi-stage loss functions for training ViTs in face recognition tasks. Experiments validate our theoretical insights, demonstrating that a carefully designed multi-stage loss outperforms single-stage alternatives.
    \item Comprehensive evaluations demonstrate LVFace's superior performance across multiple benchmarks, proving that face-specific LVMs can inherit and extend the scalability benefits of foundation vision models.
\end{itemize}

% PCO effectively guides feature cluster optimization, mitigating the challenges of FR optimization in LVMs. 
    % LVFace's architectural compatibility with mainstream VLMs and LLMs enables seamless integration into multimodal pipelines, which is a critical advantage over CNN-based systems. 
    % \item We design a cosine stage scheduler (CSS) that controls PCO's stage transitions by monitoring cosine similarity metrics. CSS tracks intra-class feature compactness, enabling robust progression through training stages without manual epoch tuning.

\section{Related Works}
\noindent
\textbf{Face Recognition.} Face recognition focuses on learning discriminative feature embeddings through the synergistic integration of backbone architectures and loss functions. Prior arts primarily follow two paradigms: softmax-based classification methods \cite{deng2019arcface, kim2022adaface, jia2023unitsface, wang2018additive, wang2018cosface, wen2021sphereface2} and metric learning approaches such as triplet loss \cite{schroff2015facenet}, tuplet loss \cite{sohn2016improved} and center loss \cite{wen2016discriminative}. While both have demonstrated promising results, they encounter insufficient discriminative power problems in large-scale/open-set scenarios, as identity numbers for face recognition dramatically grow. To address this problem, margin-based approaches such as ArcFace \cite{deng2019arcface}, CosFace\cite{wang2018cosface}, and SphereFace \cite{liu2017sphereface} introduce angular or cosine margin penalties to enhance feature discriminability. Building upon these foundations, recent methods have explored adaptive strategies: some works \cite{zhang2019p2sgrad, zhang2019adacos, baevski2018adaptive, kim2022adaface, meng2021magface} dynamically adjust margins based on sample characteristics, while others \cite{deng2021variational, fan2024epl} focus on optimizing cluster center representations. Further advancements exploring optimization directions include contrastive learning \cite{zhou2023uniface, jia2023unitsface}, inter-class regularization \cite{zhao2019regularface, duan2019uniformface}, curriculum learning \cite{huang2020curricularface}, and efficient training strategies \cite{an2022killing, an_2021_pfc_iccvw}. However, most of them have primarily been developed on CNN architectures, leaving potential for exploration within Transformer-based frameworks.

\noindent
% \textbf{Transformers.}
\textbf{Vision Transformers.}
Vision Transformers (ViTs) \cite{dosovitskiy2021imageworth16x16words} have emerged as powerful competitors to CNNs, achieving comparable performance on various vision tasks \cite{kirillov2023segment,zhang2022dino}.  
In face recognition, early ViT adaptations focused on architectural viability: FaceTransformer \cite{zhong2021facetransformerrecognition} pioneered pure-transformer frameworks, while Partial FC \cite{an2022killing} addressed scalability through sparse classifier training. Subsequent works like TransFace \cite{dan2023transfacecalibratingtransformertraining} and Part fViT \cite{sun2022part} introduced patch-level data augmentation and part-aware learning to enhance discriminability.  
However, existing ViT-based methods that directly adopt CNN-derived loss functions ($e.g.$, ArcFace \cite{deng2019arcface}) face convergence challenges during large-scale training. The inherent instability arises from ViT's unique optimization dynamics, where the interplay between high-dimensional feature distributions and the lack of local inductive biases often leads to unstable cluster formation and slow margin convergence. This limitation motivates our design of learning dynamics that explicitly stabilize ViT training through progressive optimization.

% \noindent
% \textbf{Large Vision Models for FR.}
% Large vision models are fundamental models with powerful general-purpose visual representation capabilities. CNN-based architectures, enhanced by residual connections \cite{he2016deep}, made progress in feature extraction via hierarchical local pattern learning. ViTs and their variants \cite{dosovitskiy2021imageworth16x16words,dehghani2023scaling,liu2021swin} further expanded model capacity and performance in various visual tasks. However, face recognition poses unique challenges as it demands capturing subtle facial details and discriminative features underrepresented in general-purpose training datasets \cite{deng2009imagenet, zhai2022scaling}. Existing large vision models need extensive fine-tuning for competitive face recognition performance, which is likely to be limited by computational resources. This gap highlights the urgent need for dedicated large vision models in face recognition, which could combine large-scale architecture's representation power with facial-analysis-specific features, potentially benefiting down-stream applications and researches.
\section{Preliminary}
\subsection{Problem Statement}
Open-set face recognition (FR) aims to learn a face embedding function $ f_\theta: \mathcal{I} \rightarrow \mathbb{S}^d $ on train-set $\mathcal{Y}_{\text{train}} = [\mathcal{I}_1,...,\mathcal{I}_N]$  that maps facial images $ \mathcal{I}$ to unit-norm features on a $ d $-dimensional embedding space $ \mathbb{S}^d $, such that for any testing identity $ y_p \notin \mathcal{Y}_{\text{train}} $, the decision margins maximize inter-class separability while preserving intra-class compactness between two facial identities.

\subsection{Margin-based Loss Functions} 
Recent advances in FR predominantly build upon CNNs, where refining softmax loss through discriminative margin penalties has become pivotal \cite{taigman2014deepface,liu2017sphereface, wang2018cosface, deng2019arcface}. Let $ W = [\bm{w}_1, \dots, \bm{w}_C] \in \mathbb{R}^{d \times C} $ denote the classifier weights for $ C $ training identities. Traditional softmax loss formulates FR as a closed-set multi-class classification task \cite{taigman2014deepface, cao2018vggface2}:  
% \begin{equation}
% \begin{aligned}
%     \mathcal{L}_{\text{softmax}} &= - \frac{1}{N} \sum_{i=1}^{N} \log\frac{e^{\bm{w}_{y_{i}}^\top \bm{x}_i}}{\sum_{j=1}^C e^{\bm{w}_j^\top \bm{x}_i}},  \\
%     &=\frac{1}{N} \sum_{i=1}^{N} \log(1+ \frac{\sum_{j=1,j \neq i}^C e^{\bm{w}_j^\top \bm{x}_i}}{e^{\bm{w}_{y_{i}}^\top \bm{x}_i}})
%     \label{eqsoftmax}  
% \end{aligned}  
% \end{equation}

\begin{equation}
% \begin{aligned}
    \mathcal{L}_{\text{softmax}} = - \frac{1}{N} \sum_{i=1}^{N} \log\frac{e^{\bm{w}_{y_{i}}^\top \bm{x}_i}}{\sum_{j=1}^C e^{\bm{w}_j^\top \bm{x}_i}}, 
%     &=\frac{1}{N} \sum_{i=1}^{N} \log(1+ \frac{\sum_{j=1,j \neq i}^C e^{\bm{w}_j^\top \bm{x}_i}}{e^{\bm{w}_{y_{i}}^\top \bm{x}_i}})
%     \label{eqsoftmax}  
% \end{aligned}  
\end{equation}

% \end{equation}  
where $ \bm{x}_i = f_\theta(\mathcal{I}_i) \in \mathbb{R}^{d} $ represents the facial feature of the $ i $-th image $ \mathcal{I}_i $ that belongs to $y_i$-th identity, and $ \bm{w}_j \in \mathbb{R}^{d} $ corresponds to the $ j $-th identity. While effective for closed-set scenarios ($ \mathcal{Y}_{\text{test}} \subseteq \mathcal{Y}_{\text{train}} $), 
this formulation suffers from two inherent limitations in open-set settings ($ \mathcal{Y}_{\text{test}} \cap \mathcal{Y}_{\text{train}} = \emptyset $):  
(1) traditional softmax assumes that all samples belong to known categories. Therefore, it cannot effectively process unknown-class faces and is prone to misclassifying them into known categories;
(2) it does not effectively constrain the distribution of features in the feature space, resulting in scattered intra-class features and insufficient inter-class feature distances. 
In other words, traditional softmax loss fails short to learn a compact and discriminative feature space suitable for open-set FR.

% Building on this insight,
Liu $et~al.$ \cite{liu2017sphereface} revealed that softmax-trained features exhibit intrinsic angular distributions. By reparameterizing the logit as $ \|\bm{w}_{y_{i}}\| \|\bm{x}_i\| \cos(\theta_{y_i}) $, they introduced angular margin penalties to explicitly control inter-class angular spacing. 
$ \theta_{y_i} = \arccos(\bm{w}_{y_i}^\top \bm{x}_i) $ defines the angle between the feature $ \bm{x}_i $ and its class center $ \bm{w}_{y_i} $. To isolate angular optimization, $ \bm{w}_j $ are constrained to unit norms ($ \|\bm{w}_j\|_2 = 1 $), while features are scaled to a fixed radius $ s $, yielding the normalized logit $ s \cos\theta_{y_i} $. 

This reformulation forces the network to discriminate identities purely through angular geometry:  
\begin{equation}  
    \mathcal{L}_{\text{angular}} = - \frac{1}{N} \sum_{i=1}^{N}\log\frac{e^{s \cos(\theta_{y_i})}}{e^{s \cos(\theta_{y_i})} + \sum_{j \neq y_i} e^{s \cos\theta_j}},  
\end{equation}  
where $ \theta_j = \arccos(\bm{w}_j^\top \bm{x}_i) $ is the angle between class center $\bm{w}_j$ and face feature $\bm{x}_i$. To strengthen inter-class separability, SphereFace \cite{liu2017sphereface} introduced multiplicative angular margins $ \cos(m\theta_{y_i}) $, though unstable optimization hindered its adoption. CosFace \cite{wang2018cosface} further advanced this direction by introducing additive cosine margins, which directly penalizes the cosine similarity between features and their corresponding class centers. ArcFace \cite{deng2019arcface} stabilized training via additive angular margins, and further combined the margin variants in an united framework. For simplicity, we provide the formula for sample $\bm{x}_i$ as follows:  
\begin{equation}  
    \begin{aligned}
    \mathcal{L}_{\text{uni}}(\bm{x}_i) &= -\log\frac{e^{s (\cos(m_1\theta_{y_i} + m_2)+m_3)}}{e^{s (\cos(m_1\theta_{y_i} + m_2)+m_3)} + \sum_{j \neq y_i} e^{s \cos\theta_j}}, \\
    &= \log \left( 1+ \frac{\sum_{j \neq y_i} e^{s \cos\theta_j}}{e^{s (\cos(m_1\theta_{y_i} + m_2)+m_3)}} \right).
    \end{aligned}
\end{equation}  
where $m_1$, $m_2$ and $m_3$ are the margin hyper-parameters.
For large-scale applications, Partial FC \cite{an2022killing} addressed computational bottlenecks through negative class sub-sampling during gradient updates. This approach demonstrates that training with a selected subset of class centers can achieve comparable performance to using all negative classes, while significantly reducing memory and computational overhead.

\subsection{ViT-based Face Recognition}

ViT-based face encoders typically follow the configuration of InsightFace~\cite{an_2021_pfc_iccvw}. Given an input face image $\mathcal{I} \in \mathbb{R}^{W \times W \times C}$, the framework first divides it into $N = (W/S)^2$ non-overlapping patches $\{\mathcal{I}_p^i \in \mathbb{R}^{S \times S \times C}\}_{i=1}^N$ using stride $S$. Each patch $\mathcal{I}_p^i$ is flattened into a $S^2C$-dimensional vector and linearly projected to $D$ dimensions via a trainable matrix $\mathbf{E} \in \mathbb{R}^{(S^2 C) \times D}$. These projected patch embeddings are combined with learnable positional encodings $\mathbf{E}_{\text{pos}} \in \mathbb{R}^{N \times D}$ to form the initial sequence:
\begin{equation}
\mathbf{z}_0 = [\mathcal{I}_p^1 \mathbf{E}; \cdots; \mathcal{I}_p^N \mathbf{E}] + \mathbf{E}_{\text{pos}},
\end{equation}
where the semicolon denotes row-wise concatenation. This sequence is processed through $L$ Transformer layers, each comprising multi-head self-attention (MSA) and feed-forward networks (FFN) with residual connections and layer normalization:
\begin{equation}
\begin{aligned}
\mathbf{z}'_\ell &= \text{MSA}(\text{LN}(\mathbf{z}_{\ell-1})) + \mathbf{z}_{\ell-1}, \\
\mathbf{z}_\ell &= \text{FFN}(\text{LN}(\mathbf{z}'_\ell)) + \mathbf{z}'_\ell,
\end{aligned}
\end{equation}
To preserve spatial semantics across facial regions, existing methods~\cite{an_2021_pfc_iccvw, dan2023transfacecalibratingtransformertraining} omit the dedicated \texttt{[CLS]} token used in standard ViT and instead aggregate all patch tokens from the final layer. The final feature $\bm{x}$ is obtained by concatenating all patch features $\{\mathbf{z}_L^k \in \mathbb{R}^D\}_{k=1}^N$ followed by an MLP:
\begin{equation}
\bm{x} = \text{MLP}(\text{Concat}(\mathbf{z}_L^1, \cdots, \mathbf{z}_L^N)).
\end{equation}
% This strategy explicitly maintains local-global feature interactions while enabling efficient optimization for face recognition tasks.

% \begin{figure}
%     \centering
%     \includegraphics[width=0.9\linewidth]{ICCV25Template/imgs/fig3.png}
%     \caption{Architecture. LVFace adopts a standard ViT backbone but removes the [CLS] token. Instead, it concatenates all the last hidden states as the final representation, which is then projected into a 512-dimensional face feature via an MLP.}
%     \label{fig:arch}
% \end{figure}
\section{Methodology}
In this section, we present the details of LVFace. We start with the problem statement for open-set face recognition (FR), followed by the motivation for LVFace. Then we elaborate on the Progressive Cluster Optimization (PCO), which enables LVFace to achieve state-of-the-art performance. Finally, we present the Cosine Stage Scheduler (CSS) to govern stage transitions in PCO, ensuring robust and efficient training.

\begin{figure*}
  \centering
\includegraphics[width=0.9\textwidth]{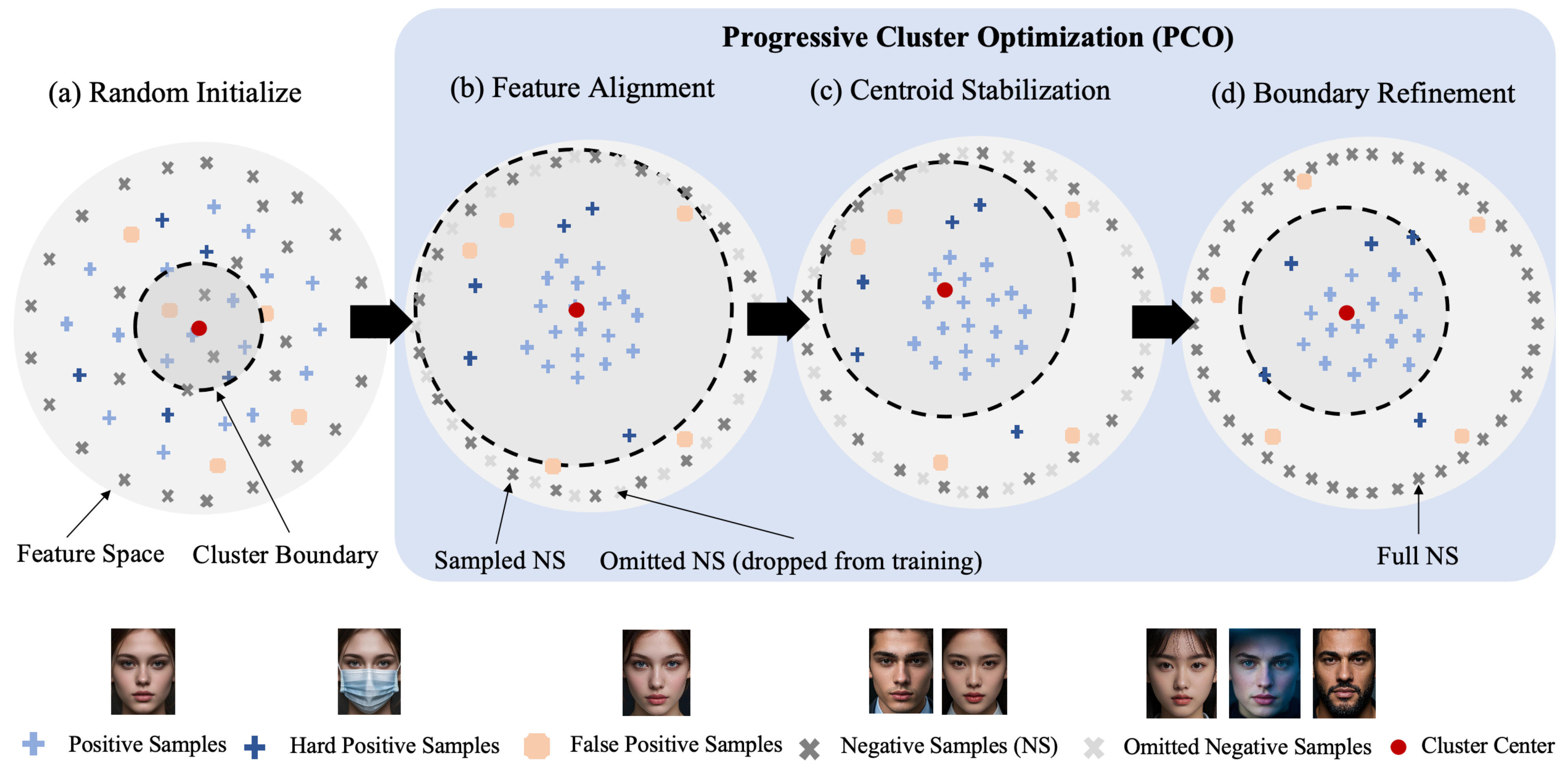}
  \caption{Overview of Progressive Cluster Optimization (PCO). We demonstrate the design philosophy of PCO in a 2-D feature space. (a) Random distribution of sample features and classifiers at the initial stage; (b) Initial feature alignment is achieved through CosFace loss and negative class sub-sampling (NCS). Positive samples aggregate at the cluster center; (c) By penalizing the feature expectation of positive samples, the training fluctuations caused by hard positive samples are gradually stabilized; (d) Disabling the NCS, unseen negative samples help to shrink cluster boundaries, achieving intra-class compactness.
  }
  \label{fig2}
\vspace{-3mm}
\end{figure*}

\subsection{Motivation}

While CNN-based methods have achieved remarkable success through extensive loss function engineering, ViT-based face recognition offers two fundamental advantages: (1) Native ViT architectures provide better compatibility with unified vision-language models (VLMs), benefiting from transformer's proven scalability in large language models (LLMs); (2) ViT's inherent parallelizability and computational efficiency enable superior representation learning on large-scale datasets.

However, Zhong $et ~al$.\cite{zhong2021facetransformerrecognition} demonstrated that ViTs face convergence challenges in FR tasks, where increasing dataset scale fails to translate into performance gains. Although Dan $et ~al$. \cite{dan2023transfacecalibratingtransformertraining} mitigated this issue through data augmentation and hard sample mining, the training paradigm requires rethinking. Inspired by the progressive training strategies in LLMs and VLMs ($e.g.$, pre-training → SFT → continual pre-training), we aim to develop a step-wise optimization strategy that conforms to natural laws of cognition, to fully unlock the potential of large vision models for face recognition.

\subsection{Progressive Cluster Optimization} 
Previous approaches typically employ a single-step optimization process, which, due to its coarse-grained learning mechanism, often leads to convergence difficulties and performance degradation when applied to Vision Transformers (ViTs). Motivated by empirical observations and inspired by \cite{huang2020curricularface}, we have developed a step-wise learning method named progressive cluster optimization (PCO). \cref{fig2} illustrates the design philosophy of PCO. PCO comprises three distinct sub-stages: \textit{feature alignment}, \textit{centroid stabilization}, and \textit{boundary refinement}. 

\noindent \textbf{Feature Alignment.}
For a specific identity/class $i$ in open-set FR scenarios, 
the initial stage typically begins with randomly initialized weights and features. This stage gradually aligns the facial features under varying conditions, such as pose and illumination, into a unified high-dimensional embedding space, as shown in \cref{fig2}(b).

However, in large-scale face datasets with millions of identities, the positive samples of the $i$-th class are vastly outnumbered by negatives, which can hinder the learning of positive patterns and the convergence of ViTs. An $et~al.$~\cite{an2022killing} showed that downsampling negatives achieves comparable performance to full-data training. To accelerate model convergence and reduce the influence of potential hard negatives ($e.g.$, those similar to positives) on the learning of positive features, we adopt a negative class sub-sampling (NCS) strategy by reducing the proportion of negative classes during training:
\begin{equation}
    S = \text{NCS}(C, r) = C*r
\end{equation}
where $S$ is the sampled negative classes, $r$ is a scalar for sub-sampling, empirically set to 0.1.
The face encoder $f_\theta$ and classifier $W$ are optimized using the CosFace loss \cite{wang2018cosface}:
\begin{equation}
\mathcal{L}_{a} = \log\left(1 + \frac{\sum_{\substack{j=0, j \neq i}}^{{S}} e^{s \cos(\theta_{j})}}{e^{s (\cos(\theta_{i}) - m)}}\right)
\label{eq:stg1}
\end{equation}

\noindent \textbf{Centroid Stabilization.}  
After the first stage, image features $\bm{x}$ are mapped to a high-dimensional embedding space $\mathbb{S}^d$ with preliminary representation capabilities. While we aim to further optimize the model by learning discriminative features from hard positives, we observe, similar to Fan $et~al.$~\cite{fan2024epl}, that some hard positives may exhibit higher similarity to negative centroids than to their own class centroid. This can mislead the classifier $\bm{w}_{i}$ during gradient updates, degrading inter-class discriminability. To address this, following \cite{fan2024epl}, we utilize the feature expectation $\bm{e}_i = \mathbb{E}(\bm{x}_i)$ as the statistical prototype for the $i$-th class in $\mathbb{S}^d$. Specifically, $\bm{e}_i$ is initialized by $\bm{x}_{i}$ and updated as:  
\begin{equation}
\bm{e}_{i}^{new} = \alpha_i \bm{e}_{i}^{old} + (1 - \alpha_i) \bm{x_i}, 
\label{eq:update_p}
\end{equation}  
where $\alpha_i$ is an adaptive coefficient defined by:  
\begin{equation}
\alpha_i = \sigma(\text{sim}(\bm{e}_i, \bm{x}_i)) = \sigma(\cos(\theta^{e}_i)),  
\label{eq:alphi_i}
\end{equation}  
with $\sigma$ as the activation function. To stabilize the positive centroid, we modify the original CosFace loss by introducing a regularization term. Specifically, we replace $\cos(\theta_*)$ with the cosine similarity $\cos(\theta^{e}_*)$ between $\bm{e}_*$ and $\bm{x}_i$, yielding:  
\begin{equation}
\mathcal{L}_{s} = \log\left(
    1 + \frac{\sum_{\substack{j=0, j \neq i}}^{S} e^{s \cos(\theta_{j})}}{e^{s (\cos(\theta_{i}) - m_1)}}
    + \frac{\sum_{\substack{j=0, j \neq i}}^{S} e^{s \cos(\theta_{j}^{e})}}{e^{s (\cos(\theta_{i}^{e}) -  m_2)}}
\right),
\label{eq:stg2}
\end{equation}  
where $m_1$ and $m_2$ are hyper-parameters controlling the cosine margin magnitude.

\noindent \textbf{Boundary Refinement.}  
While the second stage stabilizes class centroids, the learned features still lack intra-class compactness. From a decision boundary perspective, this results in overly loose cluster boundaries, limiting the model's generalization ability on unseen identities. To address this, we propose to refine the decision boundaries by introducing more negative samples, which penalize the boundaries. By disabling the NCS strategy, the model gains access to a larger pool of negatives. Crucially, the positive centroids, stabilized in the second stage, remain unaffected by the increased number of negatives, avoiding convergence issues. The loss function for this stage is defined as:  
\begin{equation}
\mathcal{L}_{r} = \log\left(
    1 + \frac{\sum_{\substack{j=0, j \neq i}}^{C} e^{s \cos(\theta_{j})}}{e^{s (\cos(\theta_{i}) - m_1)}}
    + \frac{\sum_{\substack{j=0, j \neq i}}^{C} e^{s \cos(\theta_{j}^{e})}}{e^{s (\cos(\theta_{i}^{e}) - m_2)}}
\right),
\label{eq:stg3}
\end{equation} 

\begin{figure}
    \centering
    \includegraphics[width=0.9\linewidth]{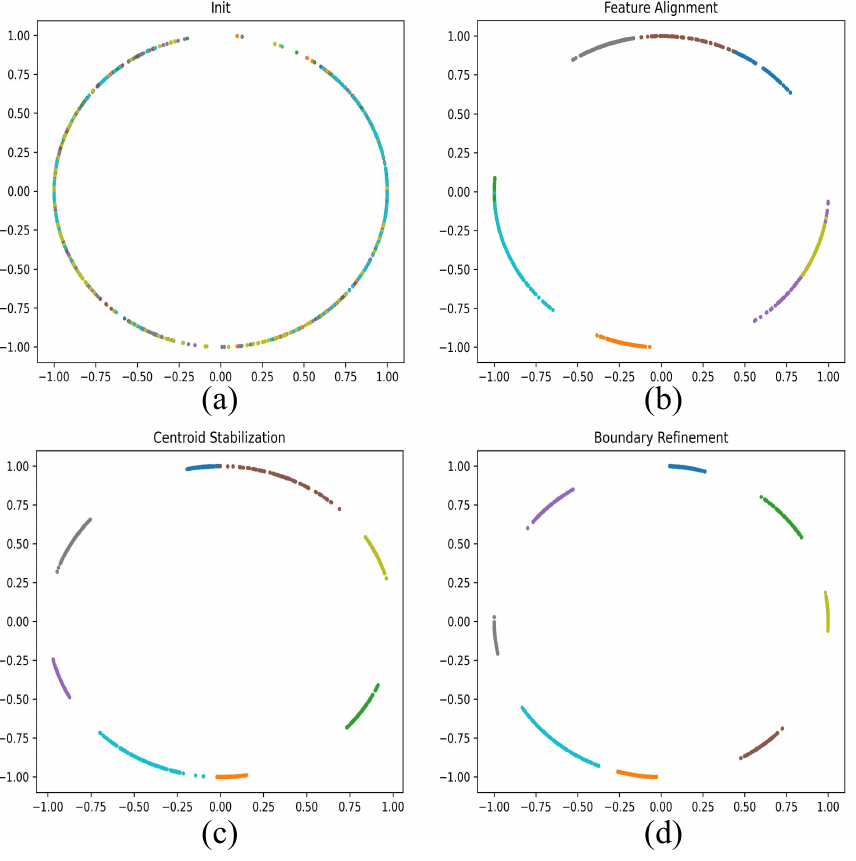}
    \caption{Feature distribution visualization across initialization and three training stages. Eight face identities are projected onto a 2D angular space (colored by class), with each point representing a single sample’s projection.}
    \label{fig:tsne}
\end{figure}

\noindent \textbf{Visualization of PCO.}
To validate the alignment between PCO's theoretical design and empirical results, we perform a t-SNE visualization of learned features $\bm{x}$, projected onto a 2D angular space where axes represent cosine distances relative to predefined reference vectors. As shown in \cref{fig:tsne}, four subplots illustrate the feature difference during optimization: \cref{fig:tsne}(a) shows chaotic cluster overlap during random initialization. In \cref{fig:tsne}(b), the \textit{Feature Alignment} stage reveals emerging class clusters with reduced intra-class dispersion, though inter-class boundaries remain ambiguous. Subsequently, \cref{fig:tsne}(c) demonstrates the \textit{Centroid Stabilization} stage, where clusters develop distinct boundaries but retain loose intra-class distributions. Finally, \cref{fig:tsne}(d) achieves compact decision boundaries through full-data refinement in the \textit{Boundary Refinement} stage. This progression empirically confirms PCO's ability to translate theoretical cluster dynamics into geometrically measurable improvements in the embedding space.

\subsection{Cosine Stage Scheduler}

To guide stage transitions in PCO, we propose a cosine stage scheduler (CSS) that monitors feature optimization progress through a similarity-based thresholding mechanism. The scheduler evaluates the optimization state by measuring the mean-square cosine similarity between sample features $\bm{x}_i = f_\theta(\mathcal{I}_i)$ and their corresponding class centroids $\bm{w}_{y_{i}}^{(t)}$ at each iteration $t$:

\begin{equation}
s^{(t)} = \frac{1}{|\mathcal{B}^{(t)}|} \sum_{\mathcal{I}_i \in \mathcal{B}^{(t)}} \|\frac{f_\theta(\mathcal{I}_i) \cdot \bm{w}_{y_i}^{(t)}}{\|f_\theta(\mathcal{I}_i)\|_2 \|\bm{w}_{y_i}^{(t)}\|_2} \|^2
\end{equation}

The optimization begins with the \textit{Feature Alignment} stage until the similarity score $s^{(t)} >= \delta_1$. Subsequently, it progresses to the \textit{Centroid Stabilization} stage until $s^{(t)} >= \delta_2$. Finally, the process enters the \textit{Boundary Refinement} stage, which continues until convergence is achieved. $\delta_1$ and $\delta_2$ are fixed thresholding scalars empirically set to 0.2 and 0.35, respectively. 

% To suppress noise introduced by mini-batch sampling, we compute a smoothed score $\tilde{s}^{(t)}$ that is maintained via exponential moving average:
% \begin{equation}
% \tilde{s}^{(t)} = (1 - \sigma) \cdot \tilde{s}^{(t-1)} + \sigma \cdot s^{(t)}
% \end{equation}
% The training process begins with negative sub-sampling and transitions to full-sample learning upon reaching $\delta_2$, creating a natural curriculum from coarse alignment to fine-grained boundary optimization. 

The pseudo code for training  LVFace is summarized in \cref{alg:pco}:
\begin{algorithm}[H]
\caption{Pseudo Code for Training LVFace}
\label{alg:pco}
\begin{algorithmic}
\REQUIRE Training set $\mathcal{Y}_{\text{train}}^C$,
          Face encoder $f_\theta$,
          Face image $\mathcal{I}$,
          Initial classifier $W$,
          Total identities $C$,
          Sub-sampling ratio $r$,
          Batch size $B$,
          Cosine stage scheduler $s^{(t)}$
\ENSURE Optimal classifier $W^{\ast}$,
         Optimal face encoder and feature $f_\theta$ \& $\bm{x}^{\ast}$

\STATE $f_\theta \sim \mathcal{N}(0, 0.01)$
\STATE $W \in \mathbb{R}^{d \times C} \sim \mathcal{U}(-1, 1)$

\STATE $\triangleright$ \textsc{Feature Alignment}
\STATE $\mathcal{Y}_{\text{train}}^{S} \gets \text{NCS}(\mathcal{Y}_{\text{train}};C, r)$ 
% \STATE For 
\FOR{batch in $\mathcal{Y}_{\text{train}}^{S}$:}
    \STATE Sample feature $\bm{x}_i = f_\theta(\mathcal{I}_i)$, $i \in [1,B]$
    \STATE Update $f_\theta, W$ with $\mathcal{L}_{a}$ \hfill // \cref{eq:stg1}
    \IF{$s^{(t)} \geq \delta_1$}
    \item BREAK; \hfill // Proceed to next stage
    \ENDIF
\ENDFOR
% \STATE $v_{\text{ind}} \gets \text{VARCALC}(\{(\bm{x}_i, y_i)\}_{i=1}^B)$

\STATE $\triangleright$ \textsc{Centroid Stabilization}
\STATE $\mathcal{Y}_{\text{train}}^{S} \gets \text{NCS}(\mathcal{Y}_{\text{train}};C, r)$ 
\FOR{batch in $\mathcal{Y}_{\text{train}}^{S}$:}
    \STATE Sample feature $\bm{x}_i = f_\theta(\mathcal{I}_i)$, $i \in [1,B]$
    \STATE Update feature expectation $\bm{e}$ \hfill // \cref{eq:update_p}
    \STATE Update $f_\theta, W$ with $\mathcal{L}_{\text{s}}$ \hfill // \cref{eq:stg2}
    \IF{$s^{(t)} \geq \delta_2$}
    \item BREAK; \hfill // Proceed to next stage
    \ENDIF
\ENDFOR

\STATE $\triangleright$ \textsc{Boundary Refinement}
\FOR{batch in $\mathcal{Y}_{\text{train}}^{C}$:}
    \STATE Sample feature $\bm{x}_i = f_\theta(\mathcal{I}_i)$, $i \in [1,C]$
    \STATE Update $f_\theta, W$ with $\mathcal{L}_{\text{r}}$ \hfill // \cref{eq:stg3}
\ENDFOR

\STATE \textbf{Return} $W^{\ast} \gets W$, $\bm{x}^{\ast} \gets f_\theta(\mathcal{I})$, $f_\theta^{\ast} \gets f_\theta$
\end{algorithmic}
\end{algorithm}

\section{Experiments}
\begin{table*}[t]
\caption{Verification accuracy (\%) on the MFR-Ongoing benchmark. Models are trained on WebFace42M \cite{zhu2022webface260m}.}
\label{tab:mfr-leaderboard}
\centering
\resizebox{\linewidth}{!}{
\small
\begin{tabular}{l|c|ccccccc|cc}
\toprule
\multirow{2}{*}{Method} & \multirow{2}{*}{Backbone} & \multicolumn{7}{c}{MFR} & \multicolumn{2}{c}{IJB-C} \\
\cmidrule(lr){3-9} \cmidrule(lr){10-11}
 & & Mask & Children & African & Caucasian & South Asian & East Asian & \textbf{MR-All} & $1e^{-5}$ & $1e^{-4}$ \\
\midrule
UniFace \cite{zhou2023uniface} & R200 & 92.43 & 93.11 & 98.14 & 98.98 & 98.84 & 90.01 & 97.92 & 96.68 & 97.91 \\
UniTSFace \cite{jia2023unitsface}  & R200 & 92.87 & 93.51 & 98.35 & 99.03 & 98.99 & 90.76 & 98.16 & 97.00 & 97.99 \\
TopoFR \cite{dan2024topofr}  & R200 & \textbf{93.96} & 93.57 & 97.97 & 98.71 & 98.98 & \textbf{92.85} & 98.13 & 97.10 & 98.01 \\
Partial FC \cite{an2022killing} & ViT-L & 90.88 & - & 98.07 & 98.81 & 98.66 & 89.97 & 97.85 & 97.23 & 98.00 \\
LVFace (Ours) & ViT-L & 93.56 & \textbf{94.31} & \textbf{98.79} & \textbf{99.26} & \textbf{99.26} & 91.02 & \textbf{98.49} & \textbf{97.25} & \textbf{98.06} \\
\bottomrule
\end{tabular}
}
\end{table*}

\subsection{Datasets} 
\noindent \textbf{Training Data:}  
To maximize model capacity, our largest variant LVFace-L is trained on WebFace42M \cite{zhu2022webface260m}, the largest publicly available high-quality face dataset, containing 42.5 million images of 2 million identities. This dataset is a refined version of WebFace260M, developed through automated quality assessment and manual verification to ensure data integrity. It features a balanced demographic distribution across age (18–65 years), ethnicity (Caucasian, Asian, African), and pose variations (±45° yaw). We further validate LVFace on Glint360K \cite{an_2021_pfc_iccvw}, a challenging dataset with 17 million images from 360,000 identities. Glint360K emphasizes real-world complexity through extreme poses (±75° yaw), heterogeneous illumination, and natural occlusions ($e.g.$, masks, hair).

\noindent \textbf{Testing Benchmarks:}  
We evaluate on three benchmarks:  
\begin{itemize}
    \item \textit{IJB-C} \cite{maze2018iarpa}: Includes 138,000 images and 11,000 video clips of 3,531 subjects, covering scenarios with extreme occlusion, low resolution, and diverse capture conditions.  
    \item \textit{IJB-B} \cite{whitelam2017ijbb}: Contains 21,800 static images and 55,000 video frames from 1,845 subjects, emphasizing cross-media (image-to-video) matching capability.  
    \item \textit{MFR-Ongoing} \cite{deng2021masked}: (ICCV-2021 Masked Face Recognition - Ongoing Challenge) The most authoritative benchmark for evaluating face recognition models' generalization performance. It includes 158,000 synthetic and real-world masked faces with 12 mask types, age-invariant verification across 10-year age gaps, balanced multi-racial cohorts under varying illuminations, and cross-quality face matching from low-resolution (16px) to high-resolution (256px).  
\end{itemize}

\subsection{Experimental Settings}
\noindent \textbf{Training Settings.}
For data preprocessing, we follow RetinaFace \cite{deng2020retinaface} to generate standardized $112 \times 112$ face crops, augmented through stochastic horizontal flipping and  normalization. LVFace's architecture comprises Vision Transformer baselines (ViT-B/ViT-L \cite{dosovitskiy2021imageworth16x16words}) as feature extractors, followed by a feature embedding MLP comprising two fully-connected layers ($512-d$ each) with intermediate BatchNorm.  
LVFace is optimized using AdamW \cite{loshchilov2017decoupled} with base learning rate 1e-3 ($\beta_1=0.9$, $\beta_2=0.999$), weight decay 0.1, and polynomial decay scheduling. We configure progressive batch size scheduling: 384 samples/batch during initial representation learning (first 60 epochs), reduced to 128 samples/batch for feature refinement (subsequent 60 epochs). Distributed training leverages automatic mixed precision (AMP) with float16/float32 casting across 64 GPUs. For hyper-parameters, we follow  
\cite{wang2018cosface} to set the feature scale $s$ to 64 and choose the angular margin $m$ at 0.4.

\noindent \textbf{Evaluation Metrics.} 
For comprehensive evaluation across the three benchmarks, we adhere to their standardized metrics: IJB-B reports True Accept Rate (TAR) at False Accept Rates (FAR=$1e^{-4}$) for verification/identification; IJB-C extends to stricter FAR=$1e^{-6}$, $1e^{-5}$ verification; MFR-Ongoing \cite{deng2021masked} as the benchmarks to test the performance of our models. The MFR-Ongoing is a comprehensive competition for evaluating FR models’ generalization performance. It contains not only the existing popular test sets, such as IJB-C, but also its own MFR benchmarks, such as Mask, Children, and Multi-Racial test sets. 
% Due to page size limitation, more training settings and experimental results are placed on Appendix.

\subsection{Results on Mainstream Benchmarks}

\subsubsection{Results on MFR-Ongoing}
The experimental results on the MFR-Ongoing benchmark demonstrate the superior generalization capability of LVFace across diverse evaluation protocols. As shown in \cref{tab:mfr-leaderboard}, LVFace achieves state-of-the-art performance on 5 out of 7 sub-tasks when trained on WebFace42M with a ViT-L backbone. While TopoFR achieves slightly better performance on the Mask subset (93.96\% vs. 93.56\%), LVFace maintains a balanced trade-off, achieving competitive results across all racial categories and securing the highest overall MR-All score of 98.49\%. Furthermore, on the IJB-C benchmark, LVFace achieves 97.25\% TAR@FAR=$1e^{-5}$ and 98.06\% TAR@FAR=$1e^{-4}$, surpassing all competitors including Partial FC (97.23\% at FAR=$1e^{-5}$), which highlights the superiority of our method in large-scale face verification tasks. Specifically, as of the submission of this work (March 2025), the proposed LVFace \textbf{ranks first} on the academic track of the MFR-Ongoing leaderboard.

\begin{table*}[t]
\caption{Verification accuracy (\%) on IJB-C and IJB-B benchmarks. GFLOPs is calculated under 112 × 112 resolution. Models are trained on Glint360K \cite{an_2021_pfc_iccvw}.}
\label{tab:ijb-bc}
\centering
\resizebox{0.9\linewidth}{!}{
\small
\begin{tabular}{l|c|c|cccc}
\toprule
 Method & Backbone & GFLOPs & IJB-C ($1e^{-6}$) & IJB-C ($1e^{-5}$) & IJB-C ($1e^{-4}$) & IJB-B ($1e^{-4}$) \\
\midrule

 ArcFace \cite{deng2019arcface} & R50 & 6.3 & 88.40 & 95.29 & 96.81 & 95.30 \\
 AdaFace \cite{kim2022adaface} & R50  & 6.3 & - & 95.58 & 96.90 & 95.66 \\
 ViT-S \cite{dan2023transfacecalibratingtransformertraining} & ViT-S & 5.7 & 88.52 & 95.24 & 96.70 & - \\
 TransFace-S \cite{dan2023transfacecalibratingtransformertraining} & ViT-S & 5.8 & 89.93 & 96.06 & \textbf{97.33} & - \\
 LVFace-S (Ours) & ViT-S & 5.7 & \textbf{90.06} & \textbf{96.52} & 97.31 & \textbf{96.14} \\
\midrule

 ArcFace \cite{deng2019arcface} & R100 & 12.1 & 88.38 & 95.38 & 96.89 & 95.69 \\
 AdaFace \cite{kim2022adaface} & R100 & 12.1 & - & 96.24 & 97.19 & 95.87 \\
 ViT-B \cite{dan2023transfacecalibratingtransformertraining} & ViT-B & 11.4 & 86.66 & 94.08 & 96.15 & - \\
 TransFace-B \cite{dan2023transfacecalibratingtransformertraining} & ViT-B & 11.5 & 88.64 & 96.18 & 97.45 & - \\
 LVFace-B (Ours) & ViT-B & 11.4 & \textbf{90.06} & \textbf{97.00} & \textbf{97.70} & \textbf{96.51} \\
\midrule

 ArcFace \cite{deng2019arcface} &R200 & 23.4 & 89.45 & 95.71 & 97.20 & 95.89 \\
 AdaFace \cite{kim2022adaface} &R200 & 23.4 & - & 95.96 & 97.33 & 96.12 \\
 ViT-L \cite{dan2023transfacecalibratingtransformertraining} & ViT-L & 25.3 & 89.69 & 95.78 & 97.13 & - \\
 TransFace-L \cite{dan2023transfacecalibratingtransformertraining} & ViT-L & 25.4 & \textbf{89.71} & 96.29 & 97.61 & - \\
 LVFace-L (Ours) & ViT-L & 25.3 & 89.51 & \textbf{97.02} & \textbf{97.66} & \textbf{96.51} \\
\bottomrule
\end{tabular}
}
\end{table*}

\subsubsection{Results on IJB-B and IJB-C}
Tab.~\ref{tab:ijb-bc} demonstrates that LVFace achieves state-of-the-art performance on IJB-C and IJB-B benchmarks across all backbone scales (ViT-S, ViT-B, ViT-L) when trained on the Glint360K dataset. At the ViT-S level, LVFace-S scores 96.52\% on IJB-C ($1e^{-5}$), outperforming both CNN-based (ArcFace R50: 95.29\%) and transformer-based competitors (TransFace-S: 96.06\%). At the ViT-B level, LVFace-B further extends its lead with 97.00\% on IJB-C ($1e^{-5}$) and 97.70\% on IJB-C ($1e^{-4}$), surpassing TransFace-B. Similarly, LVFace-L achieves 97.02\% on IJB-C ($1e^{-5}$) and 97.66\% on IJB-C ($1e^{-4}$), outperforming TransFace-L and AdaFace R200. LVFace also demonstrates consistent performance on IJB-B ($1e^{-4}$),  highlighting the robustness of the proposed PCO across diverse evaluation protocols.

\begin{table}[h]
\caption{Ablation study on the impact of network size (Tiny, Small, Base, Large) and train-sets (Glint360K, WebFace42M) on verification accuracy (\%).}
\label{tab:ablat_size}
\centering
\resizebox{\linewidth}{!}{
\begin{tabular}{l|c|cccc}
\toprule
Model & Train-set & IJB-C ($1e^{-5}$) & IJB-C ($1e^{-4}$) & IJB-B ($1e^{-4}$) \\
\midrule
LVFace-T & G360K  & 95.63 & 96.67 & 95.41 \\
LVFace-S & G360K  & 96.52 & 97.31 & 96.14 \\
LVFace-B & G360K  & 97.00 & 97.70 & 96.51 \\
LVFace-L & G360K  & 97.02 & 97.66 & 96.51 \\
LVFace-L & W42M & \textbf{97.25} & \textbf{98.06} & \textbf{96.74} \\
\bottomrule
\end{tabular}
}
\end{table}

\subsection{Ablation Studies}

We conduct extensive ablation studies to evaluate the effectiveness of LVFace and the proposed Progressive Cluster Optimization (PCO) method. Specifically, we perform three sets of ablation experiments: (1) ablation on model and training dataset scales, (2) ablation on the dependency of base loss functions, and (3) ablation on the effectiveness of each stage in the PCO strategy.

\noindent \textbf{Scalability.} As shown in \cref{tab:ablat_size}, the experiments reveal two key insights. First, on the Glint360K dataset, LVFace's performance improves as the network size increases from Tiny to Base, but the gains plateau when scaling to Large, suggesting that the dataset's limited size constrains the model's ability to fully leverage its capacity. Second, by training LVFace-L on the larger WebFace42M dataset, we achieve significant performance improvements across all benchmarks (e.g., 97.25\% on IJB-C at $1e^{-5}$ FAR). This demonstrates that large-scale datasets like WebFace42M are essential for unlocking the full potential of LVFace, highlighting the scalability and effectiveness of our method when sufficient data is available.

\begin{table}[h]
\caption{Ablation study on loss dependency. Model is trained on Glint360K with ViT-B as backbone.}
\label{tab:ablat_sens}
\centering
\resizebox{\linewidth}{!}{
\begin{tabular}{l|ccc}
\toprule
Method &  IJB-C ($1e^{-5}$) & IJB-C ($1e^{-4}$) & IJB-B ($1e^{-4}$) \\
\midrule
ArcFace Loss      & 96.11 & 97.12 & 96.01 \\
ArcFace+PCO   & 96.68 & 97.44 & 96.40 \\
CosFace Loss      & 96.15 & 97.28 & 95.99 \\
CosFace+PCO (Ours)  & \textbf{97.00} & \textbf{97.70} & \textbf{96.51} \\
\bottomrule
\end{tabular}
}
\end{table}

\noindent \textbf{Robustness.} \cref{tab:ablat_sens} demonstrates the robustness of the proposed PCO. When combined with different base loss functions (ArcFace and CosFace), PCO consistently improves performance across all benchmarks. Notably, CosFace+PCO achieves the best results, outperforming ArcFace+PCO on all metrics ($e.g.$, 97.70\% on IJB-C at $1e^{-4}$ FAR). This validates the stability of PCO and the superior compatibility of CosFace with our LVFace.

\begin{table}[h]
\caption{Ablation study of PCO on MFR-Ongoing benchmark (Accuracy\%). Experiments done on LVFace-L.}
\label{tab:ablat_pco}
\centering
\resizebox{\linewidth}{!}{
\begin{tabular}{c | c c c c c c c}
\toprule
\multirow{2}{*}{Method} & \multicolumn{7}{c}{MFR} \\
% \cline{2-8}
& \textbf{Mask} & Child & Afr & Cau & S-Asian & E-Asian & \textbf{All} \\
\midrule
ViT-L & 89.50 & 91.53 & 97.36 & 98.43 & 98.04 & 87.78 & 97.27 \\
Stage 1 & 89.99 & 91.79 & 97.73 & 98.65 & 98.37 & 87.97 & 97.52 \\
Stage 2 & 91.72 & 92.99 & 98.53 & 99.10 & 98.77 & 89.13 & 98.22 \\
Stage 3 & \textbf{93.56} & \textbf{94.31} & \textbf{98.79} & \textbf{99.26} & \textbf{99.26} & \textbf{91.02} & \textbf{98.49} \\
\bottomrule

\end{tabular}
}

\end{table}

\noindent \textbf{Effectiveness.} We show the effectiveness of our proposed PCO in \cref{tab:ablat_pco}. We observe consistent performance improvements across all stages: Stage 1 (\textit{Feature Alignment}) achieves initial gains, particularly in Mask and Child tasks; Stage 2 (\textit{Centroid Stabilization}) further enhances robustness, especially in African and Caucasian subsets; and Stage 3 (\textit{Boundary Refinement}) delivers the best results. The complete PCO boosts the All metric from 97.27\% to 98.49\%, validating its ability to address challenging face verification tasks.

\subsection{Computational Efficiency}
Our PCO introduces minimal computational overhead compared to traditional methods. While the second and third stages incorporate feature expectation penalties, the first two stages benefit from negative class sub-sampling (NCS), which reduces overall training computations through selective gradient updates. This results in comparable total training costs to conventional approaches. For inference, LVFace maintains identical latency and memory footprint to standard ViT-based models, as our method introduces no architectural modifications to the backbone network.

\section{Conclusion}
We present LVFace, a large vision model for face recognition that unlocks the full potential of ViTs through a novel Progressive Cluster Optimization (PCO) method. PCO addresses key challenges in large-scale ViT optimization by decomposing training into three progressive stages: robust feature alignment via negative class sub-sampling (NCS), centroid stabilization through feature expectation penalties, and cluster boundary refinement using full-batch training. LVFace achieves state-of-the-art performance on WebFace42M, surpassing both ViT and CNN baselines across diverse benchmarks. Our LVFace demonstrates exceptional scalability to large-scale datasets and compatibility with modern VLMs/LLMs. Our work highlights the critical role of our carefully designed optimization method in harnessing ViTs for complex visual tasks, establishing a new baseline for transformer-based face recognition systems.
{
    \small
    \bibliographystyle{ieeenat_fullname}
    \bibliography{main}
}


\end{document}